\documentclass{article}

\usepackage{preamble-arxiv-RF}
\usepackage{authblk}
\title{Trading Complexity for Sparsity \\in Random Forest Explanations} 

\author[1]{Gilles Audemard}
\author[1]{Steve Bellart}
\author[1]{Louenas Bounia}
\author[1]{Frédéric Koriche}
\author[1]{Jean-Marie Lagniez}
\author[1,2]{Pierre Marquis}
\affil[1]{CRIL, Université d'Artois \& CNRS, France}
\affil[2]{Institut Universitaire de France}

\date{email: name@cril.fr}

\begin{document}

\maketitle

\begin{abstract}
Random forests have long been considered as powerful model ensembles in machine learning. 
By training multiple decision trees, whose diversity is fostered through data and feature subsampling, 
the resulting random forest can lead to more stable and reliable predictions than a single decision tree.
This however comes at the cost of decreased interpretability: while decision trees are often easily 
interpretable, the predictions made by random forests are much more difficult to understand, as they 
involve a majority vote over hundreds of decision trees. In this paper, we examine different
types of \emph{reasons} that explain ``why'' an input instance is classified as positive or negative 
by a Boolean random forest. Notably, as an alternative to \emph{sufficient reasons} taking the form of 
prime implicants of the random forest, we introduce \emph{majoritary reasons} which are prime implicants 
of a strict majority of decision trees. For these different abductive explanations, the tractability 
of the generation problem (finding one reason) and the minimization problem (finding one shortest reason) are investigated. 
Experiments conducted on various datasets reveal the existence of a trade-off between runtime complexity and sparsity. 
Sufficient reasons - for which the identification problem is \textsf{DP}-complete - 
are slightly larger than majoritary reasons that can be generated using a simple linear-time greedy algorithm,
and significantly larger than \emph{minimal} majoritary reasons that can be approached using an anytime \textsc{Partial MaxSAT} algorithm. 
\end{abstract}

\section{Introduction}

Over the past two decades, rapid progress in statistical machine learning has led 
to the deployment of models endowed with remarkable predictive capabilities.
Yet, as the spectrum of applications using statistical learning models becomes increasingly large, 
explanations for why a model is making certain predictions are ever more critical. 
For example, in medical diagnosis, if some model predicts that an image is malignant, 
then the doctor may need to know which features in the image have led to this classification. 
Similarly, in the banking sector, if some model predicts that a customer is a fraud, 
then the banker might want to know why.
Therefore, having explanations for why certain predictions are made is essential 
for securing user confidence in machine learning technologies \cite{Miller19,Molnar19}.

This paper focuses on classifications made by \emph{random forests}, 
a popular ensemble learning method that constructs multiple randomized decision trees during the training phase, 
and predicts by taking a majority vote over the base classifiers \cite{Breiman01}. 
Since decision tree randomization is achieved by essentially coupling data subsampling (or bagging) and feature subsampling, 
random forests are fast and easy to implement, with few tuning parameters. 
Furthermore, they often make accurate and robust predictions in practice, 
even for small data samples and high-dimensional feature spaces \cite{Biau12}. 
For these reasons, random forests have been used in various applications including, 
among others, computer vision \cite{Criminisi2013}, 
crime prediction \cite{Bogomolov2014}, ecology \cite{Cutleretal2007}, 
genomics \cite{Chen2012}, and medical diagnosis \cite{Azar2014}. 

Despite their success, random forests are much less interpretable than decision trees. 
Indeed, the prediction made by a decision tree on a given data instance can be easily interpreted 
by reading the unique root-to-leaf path that covers the instance. 
Contrastingly, there is no such \emph{direct reason} in a random forest, 
since the prediction is derived from a majority vote over multiple decision trees.
So, a key issue in random forests is to infer \emph{abductive explanations}, that is, 
to explain in concise terms why a data instance is classified as positive or negative by the model ensemble.

\paragraph{Related Work.} 
Explaining random forest predictions has received increasing attention in recent years \cite{Benard2021,Choi2020,joao-ijcai21}.
Notably, in the classification setting, \cite{Choi2020,joao-ijcai21} have focused on \emph{sufficient reasons}, 
which are abductive explanations involving only relevant features \cite{DarwicheH20}. 
More specifically, if we view any random forest classifier as a Boolean function $f$, 
then a sufficient reason for classifying a data instance $\vec x$ as positive by $f$ 
is a \emph{prime implicant} $t$ of $f$ covering $\vec x$. 
By construction, removing any feature from a sufficient reason $t$ would question the fact that $t$ explains the way $\vec x$ is classified by $f$. 
Interestingly, if $f$ is described by a single decision tree, then generating a sufficient reason for any input instance $\vec x$ 
can be done in linear time. Yet, in the general case where $f$ is represented by 
an arbitrary number of decision trees, the problem of identifying a sufficient reason is {\sf DP}-complete.
Despite this intractability statement, the empirical results reported in \cite{joao-ijcai21} 
show that a MUS-based algorithm for computing sufficient reasons proves quite efficient in practice.

In addition to ``model-based'' explanations investigated in \cite{Choi2020,joao-ijcai21}, 
``model-agnostic'' explanations can be applied to random forests. 
Notably, the LIME method \cite{Lime16} extrapolates a linear threshold function 
$g$ from the behavior of the random forest $f$ around an input instance $\vec x$. 
Yet, even if a prime implicant of the linear threshold function can be easily computed, 
this explanation is \emph{not} guaranteed abductive since $g$ is only an approximation of $f$.

\paragraph{Contributions.}
In this paper, we introduce several new notions of abductive explanations:  
\emph{direct reasons} extend to the case of random forests the corresponding notion defined primarily for decision trees, 
and \emph{majority reasons} are weak forms of abductive explanations which take into account the averaging rule of random forests.
Informally, a majoritary reason for classifying a instance $\vec x$ as positive by some random forest $f$ is a prime implicant $t$ 
of a majority of decision trees in $f$ that covers $\vec x$. 
Thus, any sufficient reason is a majoritary reason, but the converse is not true. 
For these different reasons, we examine the tractability of both the generation (finding one explanation) 
and the minimization (finding one shortest explanation) problems.
To the best of our knowledge, all complexity results related to random forest explanations are new, 
if we make an exception for the intractability of generating sufficient reasons, 
which was recently established in  \cite{joao-ijcai21}. 
Notably, direct reasons and majoritary reasons can be derived in time polynomial 
in the size of the input (the instance and the random forest used to classify it).
By contrast, the identification of minimal majoritary reasons is {\sf NP}-complete, 
and the identification of minimal sufficient reasons is $\Sigma_2^p$-complete.

Based on these results, we provide algorithms for deriving random forest explanations, which open the way for an empirical comparison. 
Our experiments made on standard benchmarks show the existence of a trade-off between the runtime complexity 
of finding (possibly minimal) abductive explanations and 
the sparsity of such explanations. 
In a nutshell, majoritary reasons and minimal majoritary reasons offer interesting compromises in comparison to, 
respectively, sufficient reasons and minimal sufficient reasons. 
Indeed, the size of majoritary reasons and the computational effort required to generate them 
are generally smaller than those obtained for sufficient reasons.
Furthermore, minimal majoritary reasons outperform minimal sufficient reasons, since 
the latter are too computationally demanding.  
In fact, using an \emph{anytime} \textsc{Partial MaxSAT} solver for minimizing majoritary reasons, 
we derive sparse explanations which are typically much shorter than all other forms of abductive explanations. 
Proofs 
are reported 
in a final appendix. Additional empirical results are available on the web page
of the EXPE\textcolor{orange}{KC}TATION project: \url{http://www.cril.univ-artois.fr/expekctation/}.

\section{Preliminaries}\label{preliminaries}

For an integer $n$, let $[n] = \{1,\cdots,n\}$.
By $\mathcal F_n$ we denote the class of all Boolean functions from $\{0,1\}^n$ to $\{0,1\}$,
and  we use $X_n  = \{x_1, \cdots, x_n\}$ to denote the set of input Boolean variables.
Any Boolean vector $\vec x \in \{0,1\}^n$ is called an \emph{instance}. 
For any function $f \in \mathcal F_n$, 
an instance $\vec x \in \{0,1\}^n$ is called a \emph{positive example} of $f$ if $f(\vec x) = 1$, 
and a \emph{negative example} otherwise. 

We refer to $f$ as a propositional formula when it is described using the Boolean connectives $\land$ (conjunction), 
$\lor$ (disjunction) and $\neg$ (negation), together with the constants $1$ (true) and $0$ (false). 
As usual, a \emph{literal} $l_i$ is a variable $x_i$ or its negation $\neg x_i$, also denoted $\overline x_i$.
A \emph{term} (or \emph{monomial}) $t$ is a conjunction of literals, 
and a \emph{clause} $c$ is a disjunction of literals. 
A \emph{\dnf\ formula} is a disjunction of terms and a \emph{\cnf\ formula} 
is a conjunction of clauses. 
The set of variables occurring in a formula $f$ is denoted $\Var{f}$.
In the rest of the paper, we shall often treat instances as terms, and terms as sets of literals. 
Given an assignment $\vec z \in \{0,1\}^n$, the corresponding term is defined as
\begin{align*}
t_{\vec z} = \bigwedge_{i=1}^n x_i^{z_i} \mbox{ where } x_i^0 = \overline x_i \mbox{ and } x_i^1 = x_i
\end{align*}
A term $t$ \emph{covers} an assignment $\vec z$ if $t \subseteq t_{\vec z}$. 
An \emph{implicant} of a Boolean function $f$
is a term that implies $f$, that is, a term $t$ such that $f(\vec z) = 1$ for every assignment $\vec z$ covered by $t$. 
A \emph{prime implicant} of $f$ is an implicant $t$ of $f$ such that no proper subset of $t$ is an implicant of $f$.

With these basic notions in hand, a (Boolean) \emph{decision tree} on $X_n$ is a binary tree $T$, 
each of whose internal nodes is labeled with one of $n$ input variables, 
and whose leaves are labeled $0$ or $1$. 
Every variable is supposed (w.l.o.g.) to occur at most once on any root-to-leaf path (read-once property).  
The value $T(\vec x) \in \{0,1\}$ of $T$ on an input instance $\vec x$ is given by the label of the
leaf reached from the root as follows: at each node go to the left or right child depending 
on whether the input value of the corresponding variable is $0$ or $1$, 
respectively. A (Boolean) \emph{random forest} on $X_n$ is an ensemble $F = \{T_1,\cdots,T_m\}$, where each $T_i$ $(i \in [m])$ 
is a decision tree on $X_n$, and such that the value $F(\vec x) \in \{0,1\}$ on an input instance $\vec x$ is given by 
\begin{align*}
  F(\vec x) = 
    \begin{cases}
      1 & \mbox{ if } \frac{1}{m}\sum_{i=1}^m T_i(\vec x)  > \frac{1}{2} \\
      0 & \mbox{ otherwise.}
    \end{cases}
\end{align*}
The size of $F$ is given by $\size{F} = \sum_{i=1}^m \size{T_i}$,
where $\size{T_i}$ is the number of nodes occurring in $T_i$. The class of decision trees on 
$X_n$ is denoted $\dt_n$, and the class of random forests with at most $m$ decision trees (with $m \geq 1$)
over $\dt_n$ is denoted $\rf_{n,m}$. $\rf_n$ is the union of all $\rf_{n,m}$ for $m \in \mathbb N$.
   
\begin{figure}[t]
\centering
\scalebox{0.9}{
      \begin{tikzpicture}[scale=0.7, roundnode/.style={circle, draw=gray!60, fill=gray!5, very thick, minimum size=7mm},
      squarednode/.style={rectangle, draw=red!60, fill=red!5, very thick, minimum size=5mm}]
        \node[roundnode](root) at (2,7){$x_4$};
        \node at (0.5,7){$T_1$};
        \node[squarednode](n1) at (1,5){$0$};
        \node[roundnode](n2) at (3,5){$x_2$};
        \node[squarednode](n21) at (2,3){$1$};
        \node[roundnode](n22) at (4,3){$x_3$};  
        \node[squarednode](n221) at (3,1){$0$};
        \node[roundnode](n222) at (5,1){$x_1$};   
        \node[squarednode](n2221) at (4,-1){$0$};
        \node[squarednode](n2222) at (6,-1){$1$};
        \draw[dashed] (root) -- (n1);
        \draw(root) -- (n2);
        \draw[dashed] (n2) -- (n21);
        \draw(n2) -- (n22);    
        \draw[dashed] (n22) -- (n221);
        \draw(n22) -- (n222);  
        \draw[dashed] (n222) -- (n2221);
        \draw(n222) -- (n2222);
                
        \node[roundnode](root2) at (9,7){$x_2$};
        \node at (7.5,7){$T_2$};
        \node[roundnode](n2-1) at (8,5){$x_1$};  
        \node[squarednode](n2-2) at (10,5){$1$};
        \node[squarednode](n2-11) at (7,3){$0$};
        \node[roundnode](n2-12) at (9,3){$x_4$};  
        \node[squarednode](n2-121) at (8,1){$0$};
        \node[squarednode](n2-122) at (10,1){$1$};
        \draw[dashed] (root2) -- (n2-1);
        \draw(root2) -- (n2-2);
        \draw[dashed] (n2-1) -- (n2-11);
        \draw(n2-1) -- (n2-12);    
        \draw[dashed] (n2-12) -- (n2-121);
        \draw(n2-12) -- (n2-122);  
        
        \node[roundnode](root3) at (16,7){$x_3$};
        \node at (14.5,7){$T_3$};
        \node[roundnode](n3-1) at (14,5){$x_2$};  
        \node[roundnode](n3-2) at (17,5){$x_2$};  
        \node[roundnode](n3-11) at (13,3){$x_1$};
        \node[roundnode](n3-12) at (15,3){$x_4$};   
        \node[squarednode](n3-121) at (14.5,1){$0$};
        \node[roundnode](n3-122) at (16,1){$x_1$};  
        \node[squarednode](n3-1221) at (15,-1){$0$};
        \node[squarednode](n3-1222) at (17,-1){$1$};
        \node[squarednode](n3-21) at (16.5,3){$0$};
        \node[roundnode](n3-22) at (18,3){$x_4$};     
        \node[squarednode](n3-111) at (12,1){$0$};
        \node[squarednode](n3-112) at (13.5,1){$1$};
        \node[squarednode](n3-221) at (17.5,1){$0$};
        \node[squarednode](n3-222) at (19,1){$1$};
        \draw[dashed] (root3) -- (n3-1);
        \draw(root3) -- (n3-2);
        \draw[dashed] (n3-1) -- (n3-11);
        \draw(n3-1) -- (n3-12);    
        \draw[dashed] (n3-2) -- (n3-21);
        \draw(n3-2) -- (n3-22); 
        \draw[dashed] (n3-11) -- (n3-111);
        \draw(n3-11) -- (n3-112);  
        \draw[dashed] (n3-22) -- (n3-221);
        \draw(n3-22) -- (n3-222);
        \draw[dashed] (n3-12) -- (n3-121);
        \draw(n3-12) -- (n3-122);
        \draw[dashed] (n3-122) -- (n3-1221);
        \draw(n3-122) -- (n3-1222);
      \end{tikzpicture}
 }
    \caption{A random forest $F = \{T_1, T_2, T_3\}$ for recognizing {\it Cattleya} orchids. The left (resp. right) child of any decision node labelled by $x_i$ corresponds to 
    the assignment of $x_i$ to $0$ (resp. $1$).\label{fig:orchids}}
    \end{figure}
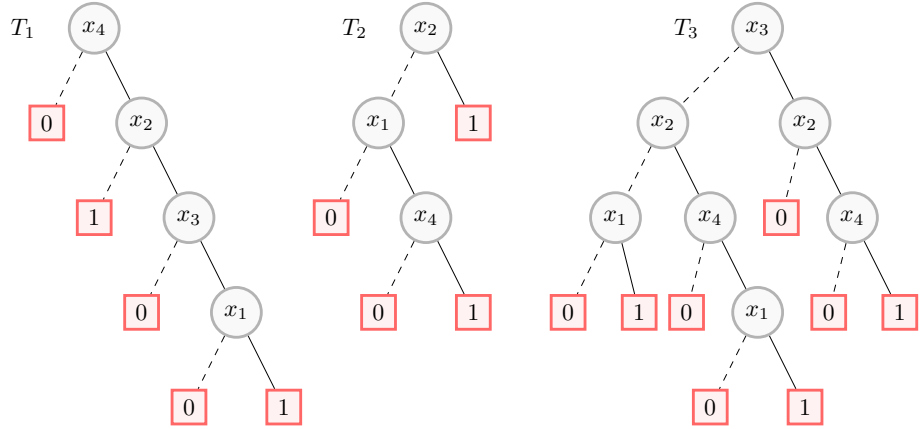

\begin{example}\label{running-ex}
The random forest $F = \{T_1, T_2, T_3\}$ in Figure \ref{fig:orchids} is composed of three decision trees. It separates {\it Cattleya} orchids from other orchids using the following features:
$x_1$: ``has fragrant flowers'', $x_2$: ``has one or two leaves'', $x_3$: ``has large flowers'', and $x_4$: ``is sympodial''.
\end{example}

It is well-known that any decision tree $T$ can be transformed into its negation $\neg T \in \dt_n$, 
by simply reverting the label of leaves. Negating a random forest can also be achieved in polynomial time:  

\begin{proposition}\label{prop:transfoRF}~
  There exists a linear-time algorithm that computes a random forest $\neg F \in \rf_{n,m}$ equivalent to
  the negation of a given random forest $F \in \rf_{n,m}$.
\end{proposition}

Another important property of decision trees is that any $T \in \dt_n$ can be transformed in linear time 
into an equivalent disjunction of terms $\dnf(T)$, where each term coincides with a $1$-path (i.e., a path from the root to a leaf labeled with $1$), 
or a conjunction of clauses $\cnf(T)$, where each clause is the negation of term describing a $0$-path. 
When switching to random forests, the picture is quite different:

\begin{proposition}\label{prop:CNF2RF} 
  Any  \cnf\ or \dnf\ formula can be converted in linear time into an equivalent random forest, but there is no polynomial-space translation from \rf\ to \cnf\ or to \dnf.
\end{proposition}

\section{Random Forest Explanations}\label{sec:explanations}

The key focus of this study is to explain \emph{why} a given (Boolean) random forest classifies 
some incoming data instance as positive or negative.
This calls for a notion of 
abductive explanation\footnote{Unlike \cite{IgnatievNM19}, 
we do not require those explanations to be minimal w.r.t. set inclusion, in order to keep the concept 
distinct (and actually more general) then the one of sufficient reasons.}. 
Formally, given a Boolean function $f \in \mathcal F_n$ and an instance $\vec x \in \{0, 1\}^n$, 
an {\em abductive explanation} for $\vec x$ given $f$ is an implicant $t$ of $f$ (resp. $\neg f$) if $f(\vec x) = 1$ (resp. $f(\vec x) = 0$) that covers $\vec x$. 
An abductive explanation $t$ for $\vec x$ given $f$ always exists, since $t = t_{\vec x}$ is such a (trivial) explanation. 
So, in the rest of this section, we shall mainly concentrate on \emph{sparse} forms of abductive explanations.

Before delving into details, it is worth mentioning that if $f$ is represented by a random forest then, without loss of generality, 
we can focus on the case where $\vec x$ is a positive example of $f$, because $\neg f$ can be computed in linear time (by Proposition~\ref{prop:transfoRF}). 
Nevertheless, for the sake of clarity, we shall consider both cases $f(\vec x) = 1$ and $f(\vec x) = 0$ in our definitions.

\subsection{Direct Reasons}

For a decision tree $T \in \dt_n$ and a data instance $\vec x \in \{0, 1\}^n$, the \emph{direct reason} of $\vec x$ given $T$ 
is the term $t_{\vec x}^T$ corresponding to the unique root-to-leaf path of $T$ that covers $\vec x$. 
We can extend this simple form of abductive explanation to random forests as follows: 

\begin{definition}
Let $F = \{T_1, \ldots, T_m\}$ be a random forest in $\rf_{n,m}$, and $\vec x \in \{0,1\}^n$ be an instance.
Then, the {\em direct reason} for $\vec x$ given $F$ is the term $t_{\vec x}^F$ defined by
\begin{align*}
t_{\vec x}^F = 
\begin{cases}
  \bigwedge_{T_i \in F : T_i(\vec x) = 1} t_{\vec x}^{T_i} & \mbox{ if } F(\vec x) = 1\\
  \bigwedge_{T_i \in F : T_i(\vec x) = 0} t_{\vec x}^{T_i} & \mbox{ if } F(\vec x) = 0
\end{cases}
\end{align*}
\end{definition}

By construction, $t_{\vec x}^F$ is an abductive explanation which can be computed in $\mathcal O(\size F)$ time.

\begin{example}
Considering Example \ref{running-ex} again, the instance $\vec x = (1, 1, 1, 1)$ 
is recognized as a {\it Cattleya} orchid, since $F(\vec x) = 1$.
The direct reason for $\vec x$ given $F$ is $t_{\vec x}^F = x_1 \land x_2 \land x_3 \land x_4$. 
It coincides with $t_{\vec x}$. 
Consider now the instance $\vec x' = (0, 1, 0, 0)$; 
it is not recognized as a {\it Cattleya} orchid, since $F(\vec x) = 0$.
The direct reason for $\vec x'$ given $F$ is $t_{\vec x'}^F = x_2 \land \overline{x}_3 \land \overline{x}_4$. It is a better abductive explanation than $t_{\vec x'}$ itself
since it does not contain 
$\overline{x}_1$, which is locally irrelevant.
\end{example}

\subsection{Sufficient Reasons}

Another valuable notion of abductive explanation is the one of {\em sufficient reason}\footnote{Sufficient reasons are also known as prime-implicant explanations \cite{ShihCD18}.}, 
defined for any Boolean classifier \cite{DarwicheH20}. In the setting of random forests, such explanations can be defined as follows: 

\begin{definition} 
  Let $F \in \rf_{n}$ be a random forest and $\vec x \in \{0,1\}^n$ be an instance.
  A \emph{sufficient reason} for $\vec x$ given $F$ is a prime implicant $t$ of $F$ (resp. $\neg F$) if $F(\vec x) = 1$ (resp. $F(\vec x) = 0$) that covers $\vec x$.
\end{definition}

\begin{example} 
For our running example, $x_2 \wedge x_3 \wedge x_4$ and $x_1 \wedge x_4$ 
are the sufficient reasons for $\vec x$ given $F$. $\overline{x}_4$ and $\overline{x}_1 \wedge \overline{x}_3$ are
the sufficient reasons for $\vec x'$ given $F$.
\end{example}

Unlike arbitrary abductive explanations, all features occurring in a sufficient reason $t$ are \emph{relevant}. Indeed, 
removing any literal from $t$ would question the fact that $t$ implies $F$. To this point, 
the direct reason $t_{\vec x}^F$ for $\vec x$ given $F$ may contain arbitrarily many more features 
than a sufficient reason for $\vec x$ given $F$,
since this was already shown in the case where $F$ consists in a single decision tree  \cite{DBLP:journals/corr/abs-2010-11034}.

The problem of identifying a sufficient reason $t$ for an input instance $\vec x \in \{0,1\}^n$ with respect to a given random 
forest $F \in \rf_n$, has recently been shown {\sf DP}-complete \cite{joao-ijcai21}. In fact, even the apparently simple task 
of \emph{checking} whether $t$ is an implicant of $F$ is already hard: 

\begin{proposition} \label{prop:complexityimplicanttestRF}
  Let $F$ be a random forest in $\rf_{n}$ and $t$ be a term over $X_n$. 
  Then, deciding whether $t$ is an implicant of $F$ is {\sf coNP}-complete.
\end{proposition}

The above result is in stark contrast with the computational complexity of checking whether a term $t$ is an implicant of a decision tree $T$.
This task can be solved in polynomial time, using the fact that $T$ can be converted (in linear time) into its clausal form 
$\cnf(T)$, together with the fact that testing whether $t$ implies $\cnf(T)$ can be done in $\mathcal O(\size{T})$ time. 
That mentioned, in the case of random forests, the implicant test can be achieved via a call
to a {\sc SAT} oracle:

\begin{proposition} \label{prop:implicanttestRF}
Let $F = \{T_1, \ldots, T_m\}$ be a random forest of  $\rf_{n, m}$, and $t$ be a (satisfiable) term over $X_n$. 
Let $H$ be the \cnf\ formula 
\begin{align*}
\{(\overline y_i \vee c) : i \in [m], c \in \cnf(\neg T_i) \} \cup \cnf \left(\sum_{i = 1}^m y_i > \frac{m}{2} \right)
\end{align*}
where $\{y_1, \ldots, y_m\}$ are fresh variables and $\cnf\left(\sum_{i = 1}^m y_i > \frac{m}{2}\right)$ is a \cnf\ encoding of the
cardinality contraint $\sum_{i = 1}^m y_i > \frac{m}{2}$. Then, $t$ is an implicant of $F$ if and only if $H \land t$ is unsatisfiable.
\end{proposition}

Based on such an encoding, 
the sufficient reasons for an instance $\vec x$ given a random forest $F$ can be characterized in terms of
MUS (minimal unsatisfiable subsets), as suggested in \cite{joao-ijcai21}. 
This characterization is useful because many {\sc SAT}-based algorithms 
for computing a MUS (or even all MUSes) of a \cnf\ formula have been pointed out for the past decade
\cite{AudemardLS13,LiffitonPMM16,MarquesSilvaJM17}, and hence, 
one can take advantage of them for computing sufficient reasons. 

Going one step further, a natural way for improving the clarity of sufficient reasons is to focus on those of minimal size.
Specifically, given $F \in \rf_n$ and $\vec x \in \{0,1\}^n$, a \emph{minimal sufficient reason} for $\vec x$ with respect to $F$
is a sufficient reason for $\vec x$ given $F$ of minimal size.\footnote{Minimal sufficient reasons should not to be confused with \emph{minimum-cardinality explanations}  \cite{ShihCD18}, where the minimality condition bears on the features set to $1$ in the data instance $\vec x$.}

\begin{example} For our running example, $x_1 \wedge x_4$ is the unique minimal sufficient reason for $\vec x$ given $F$, and $\overline{x}_4$ 
is the unique minimal reason for $\vec x'$ given $F$.
\end{example}

As a by-product of the characterization of a sufficient reason in terms of MUS \cite{joao-ijcai21}, 
a minimal sufficient reason for $\vec x$ given $f$ can be viewed as a \emph{minimal} MUS. Thus, we can
exploit algorithms for computing minimal MUSes (see e.g., \cite{IgnatievPLM15}) 
in order to derive minimal sufficient reasons. 
However, deriving a minimal sufficient reason is computationally harder than deriving a sufficient reason:

\begin{proposition}\label{prop:complexityminimalMUS}
    Let $F \in \rf_n$, $\vec x \in \{0,1\}^n$, and $k \in \mathbb N$. 
    Then, deciding whether there exists a minimal sufficient reason $t$ for $\vec x$ given $F$ containing at most $k$ features is $\Sigma_2^p$-complete.
\end{proposition}

\subsection{Majoritary Reasons}

Based on the above considerations, a natural question arises: does there exist a middle ground between direct reasons, 
which main contain many irrelevant features but are easy to calculate, and sufficient reasons, which only contain relevant 
features but are potentially much harder to generate?  Inspired by the way prime implicants can be computed 
when dealing with decision trees, we can reply in the affirmative using the notion of \emph{majoritary reasons}, defined as follows.

\begin{definition}
Let $F = \{T_1, \ldots, T_m\}$ be a random forest in $\rf_{n,m}$ and $\vec x \in \{0, 1\}^n$ be an instance.
Then, a \emph{majoritary reason} for $\vec x$ given $F$ is a term $t$ covering $\vec x$, such that $t$ 
is an implicant of at least $\lfloor \frac{m}{2} \rfloor +1$ decision trees $T_i$ (resp. $\neg T_i$) if $F(\vec x) = 1$ (resp. $F(\vec x) = 0$), 
and for every $l \in t$, $t \setminus \{l\}$ does not satisfy this last condition.
 \end{definition}

\begin{example} For our running example, $\vec x$ has three majoritary reasons given $F$: $x_1 \land x_2 \land x_4$, $x_1 \land x_3 \land x_4$, and 
$x_2 \land x_3 \land x_4$. Those reasons are better than $t_{\vec x}^F$ in the sense that they are shorter than this direct reason.
$\vec x'$ has three majoritary reasons given $F$: $\overline{x}_1 \land \overline{x}_4$, $x_2 \land \overline{x}_4$, and 
$\overline{x}_1 \land x_2 \land \overline{x}_3$. 
Each of the two majoritary reasons $x_1 \land x_2 \land x_4$ and $x_1 \land x_3 \land x_4$ for $\vec x$ given $F$ contains
an irrelevant literal for the task of classifying $\vec x$ using $F$ since $x_1 \land x_4$ is a sufficient reason for $\vec x$ given $F$.
Similarly, each majoritary reason for $\vec x'$ given $F$ contains an irrelevant literal for the task of classifying $\vec x'$ using $F$.
\end{example}

As the previous example illustrates it, the notions of majoritary reasons and of sufficient reasons do not coincide in general. Indeed, 
a sufficient reason $t$ is a prime implicant (covering $\vec x$) of the forest $F$, while a majoritary reason $t'$ 
is an implicant (covering $\vec x$) of a strict majority of decision trees in the forest $F$ satisfying the 
additional condition that $t'$ is a prime implicant of at least one of these decision trees. Viewing majoritary 
reasons as ``weak'' forms of sufficient reasons, they can include irrelevant features:

\begin{proposition} \label{prop:majoritary}
 Let $F = \{T_1, \ldots, T_m\}$ be a random forest of  $\rf_{n, m}$ and $\vec x \in \{0,1\}^n$ such that $F(\vec x) = 1$. Unless $m < 3$,
 it can be the case that every majoritary reason for $\vec x$ given $F$ contains arbitrarily many more features 
 than any sufficient reason for $\vec x$ given $F$.
\end{proposition}

What makes majoritary reasons valuable is that they are abductive and can be generated in linear time. 
The evidence that any majoritary reason $t$ for $\vec x$ given $F$ is an abductive explanation for $\vec x$ given $F$ 
comes directly from the fact that if $t$ implies a majority of decision trees in $F$, then it is an implicant of $F$ 
(note that the converse implication does not hold in general). 

The tractability of generating majoritary reasons lies in 
the fact that they can be found using a simple greedy algorithm. For the case where $F(\vec x) = 1$,  
start with $t = t_{\vec x}$, and iterate over the literals $l$ of $t$ by checking whether $t$ deprived of $l$ 
is an implicant of at least $\lfloor \frac{m}{2} \rfloor +1$ decision trees of $F$. 
If so, remove $l$ from $t$ and proceed to the next literal.
Once all literals in $t_{\vec{x}}$ have been examined, the final term $t$ is by construction an implicant of a strict majority 
of decision trees in $F$, such that removing any literal from it would lead to a term that is no longer an implicant of this majority. 
So, $t$ is by construction a majoritary reason. The case where $F(\vec x) = 0$ is similar, by simply replacing each $T_i$ with its negation in $F$.
This greedy algorithm runs in $\mathcal O(n\size{F})$ time, using the fact that, on each iteration, checking whether 
$t$ is an implicant of $T_i$ (for each $i \in [m]$) can be done in $\mathcal O(\size{T_i})$ time.

By analogy with minimal sufficient reasons, a natural way of improving the quality of majoritary reasons 
is to seek for shortest ones. Let $F \in \rf_n$ be a random forest and $\vec x \in \{0, 1\}^n$ be an instance.
Then, a {\em minimal majoritary reason} for $\vec x$ given $F$ is a minimal-size majoritary reason for $\vec x$ given $F$.

\begin{example} For our running example, the three majoritary reasons for $\vec x$ given $F$ 
are its minimal majoritary reasons. Contrastingly, among the majoritary reasons for $\vec x'$ given $F$, only 
$\overline{x}_1 \land \overline{x}_4$ and $x_2 \land \overline{x}_4$ are minimal majoritary reasons.
\end{example}

Unsurprisingly, the optimization task for majoritary reasons is more demanding than the generation task. Yet, 
minimal majoritary reasons are easier to find than minimal sufficient reasons. Specifically:  

\begin{proposition}\label{prop:minimal majoritaryreasonRF}
  Let $F \in \rf_n$, $\vec x \in \{0,1\}^n$, and $k \in \mathbb N$. 
  Then, deciding whether there exists a minimal majoritary reason $t$ for $\vec x$ given $F$ containing at most $k$ features is {\sf NP}-complete.
\end{proposition}
  

A common approach for handling {\sf NP}-optimization problems is to rely on modern constraint solvers.
From this perspective, recall that a \textsc{Partial MaxSAT} problem consists of a pair  $(C_{\mathrm{soft}}, C_{\mathrm{hard}})$
where $C_{\mathrm{soft}}$ and  $C_{\mathrm{hard}}$ are (finite) sets of clauses. 
The goal is to find a Boolean assignment that maximizes the number of 
clauses $c$ in $C_{\mathrm{soft}}$ that are satisfied, while satisfying all clauses in $C_{\mathrm{hard}}$.

\begin{proposition}\label{prop:minimal-weightoptim}
    Let $F \in \rf_{n,m}$ and $\vec x \in \{0,1\}^n$ be an instance such that $F(\vec x) = 1$. 
    Let $(C_{\mathrm{soft}}, C_{\mathrm{hard}})$ be an instance of the \textsc{Partial MaxSAT} problem such that:
    \begin{align*}
        C_{\mathrm{soft}} &= \{\overline x_i  : x_i \in t_{\vec x}\} \cup \{x_i  : \overline x_i \in t_{\vec x}\} \\
        C_{\mathrm{hard}} &= \{(\overline y_i \vee c_{\mid \vec x}) : i \in [m], c \in \cnf(T_i) \} \cup \cnf\left(\sum_{i = 1}^m y_i > \frac{m}{2}\right)
    \end{align*}
    where $c_{\mid \vec x} = c \cap t_{\vec x}$ is the restriction of $c$ to the literals in $t_{\vec x}$, 
    $\{y_1, \ldots, y_m\}$ are fresh variables and $\cnf(\sum_{i = 1}^m y_i > \frac{m}{2})$ is a \cnf\ encoding of the
    contraint $\sum_{i = 1}^m y_i > \frac{m}{2}$. 
    The intersection of $t_{\vec x}$ with  $t_{\vec z^*}$, where $\vec z^*$ is an 
    optimal solution of $(C_{\mathrm{soft}}, C_{\mathrm{hard}})$, is a minimal majoritary reason for $\vec x$ given $F$.
\end{proposition}

Clearly, in the case where $F(\vec x) = 0$, it is enough to consider the same instance of \textsc{Partial MaxSAT} as above, except that
$C_{\mathrm{hard}} = \{(\overline y_i \vee c_{\mid \vec x}) : i \in [m], c \in \cnf(\neg T_i) \} \cup \cnf(\sum_{i = 1}^m y_i > \frac{m}{2})$. 

Thanks to this characterization result, one can leverage the numerous algorithms that have been developed so far for \textsc{Partial MaxSAT} 
(see e.g. \cite{AnsoteguiBL13,MorgadoIM14,NarodytskaB14,SaikkoBJ16}) 
in order to compute minimal majoritary reasons. 
We took advantage of it to achieve some of the experiments reported in Section \ref{experiments}.

\section{Experiments}\label{experiments}



\paragraph{Empirical setting.}

The empirical protocol was as follows.  We have considered 15
datasets, which are standard benchmarks from the well-known
repositories Kaggle (\url{www.kaggle.com}), OpenML
(\url{www.openml.org}), and UCI (\url{archive.ics.uci.edu/ml/}). 
These datasets are \textit{compas}, \textit{ placement}, \textit{recidivism}, \textit{adult}, \textit{ad\_data}, \textit{mnist38}, \textit{mnist49}, 
\textit{gisette}, \textit{dexter}, \textit{dorothea}, \textit{farm-ads}, \textit{higgs\_boson}, \textit{christine}, \textit{gina}, and \textit{bank}.
\textit{mnist38} and \textit{mnist49} are subsets of the \textit{mnist} dataset, restricted to the instances of 3 and 8 (resp. 4 and 9) digits.
Additional information
about the datasets (especially the numbers and types of features, the number of instances), 
and about the random forests that have been trained (especially, the number of Boolean features used, 
the number of trees, the depth of the trees, the mean accuracy) 
can be found at \url{http://www.cril.univ-artois.fr/expekctation/}.
We used only datasets for binary classification, which is a very common kind of dataset.  Categorical
features have been treated as arbitrary numbers (the scale is
nominal).  As to numeric features, no data preprocessing has taken
place: these features have been binarized on-the-fly by the random
forest learning algorithm that has been used.


For every benchmark $b$, a $10$-fold cross validation process has been
achieved.  Namely, a set of $10$ random forest $F_b$ have been
computed and evaluated from the labelled instances of $b$, partitioned
into $10$ parts.  One part was used as the test set and the remaining
$9$ parts as the training set for generating a random forest. The
classification performance for $b$ was measured as the mean accuracy
obtained over the $10$ random forests generated from $b$.
As to the random forest learner, we have used the implementation
provided by the \textrm{Scikit-Learn} \cite{scikit-learn} library in
his version 0.23.2.  The maximal depth of any decision tree in a forest has been bounded at 8.
All other hyper-parameters of the learning algorithm
have been set to their default value except the number of trees. We
made some preliminary tests for tuning this parameter in order to 
ensure that the accuracy is good enough.
%
For each benchmark $b$, each random forest $F$, and a subset of 
25 instances $\vec x$ picked up at random in the corresponding
test set (leading to 250 instances per dataset) we have run the
algorithms described in Section \ref{sec:explanations} for deriving
the direct reason for $\vec x$ given $F$, a sufficient reason for
$\vec x$ given $F$, a majoritary reason $\vec x$ given $F$, a
minimal majoritary reason for $\vec x$ given $F$, and a minimal sufficient reason for $\vec x$ given $F$.
%
%

For computing sufficient reasons and minimal majoritary
reasons, we took advantage of the Pysat library~\cite{IgnatievMM18} (version 
0.1.6.dev15) which provides the implementation of the RC2
\textsc{Partial MaxSAT} solver and an interface to MUSER~\cite{muser}.
When deriving majoritary reasons, we picked up uniformly at random
50 permutations of the literals describing the instance and tried to eliminate
those literals (within the greedy algorithm) following the 
ordering corresponding to the permutation. As a majoritary reason for the instance, we kept 
a smallest reason among those that have been derived (of course, the corresponding 
computation time that has been measured is the cumulated time over the 50 tries).
Sufficient reasons have been computed as MUSes, as explained before.

We also derived a ``LIME explanation'' for each
instance. Such an explanation has been generated thanks to the
following approach.  For any $\vec x$ under consideration, one first used 
LIME \cite{Lime16} to generate an associated linear model
$\vec w_{\vec x}$ where $\vec w_{\vec x} \in \mathbb{R}^n$.  This
linear model $\vec w_{\vec x}$ classifies any instance $\vec x'$ as a
positive instance if and only if $\vec w_{\vec x} \cdot \vec x' > 0$.
Furthermore, $\vec w_{\vec x}$ classifies the instance to be explained
$\vec x$ in the same way as the black box model considered at start
(in our case, the random forest $F$). We ran the LIME
implementation linked to \cite{Lime16} in its latest version.
Interestingly, a minimal sufficient reason $t$ for $\vec x$ given
$\vec w_{\vec x}$ can be generated in polynomial time from
$\vec w_{\vec x}$. We call it a LIME explanation for $\vec x$. The
computation of $t$ is as follows.  If $\vec x$ is classified
positively by $\vec w_{\vec x}$, in order to derive $t$, it is enough
to sum in a decreasing way the positive weights $w_i$ occurring in
$\vec w_{\vec x}$ until this sum exceeds the sum of the opposites of
all the negative weights occurring in $\vec w_{\vec x}$. The term $t$
composed of the variables $x_i$ corresponding to the positive weights
that have been selected is by construction a minimal sufficient reason for
$\vec x$ given $\vec w_{\vec x}$ since for every $\vec x'$ covered by
$t$, the inequation $\vec w_{\vec x} \cdot \vec x' > 0$ necessarily
holds; indeed, it holds in the worst situation where all the variables
associated with a positive weight in $\vec w_{\vec x}$ and not
belonging to $t$ are set to $0$, whilst all the variables associated
with a negative weight in $\vec w_{\vec x}$ are set to $1$. Similarly,
if $\vec x$ is classified negatively by $\vec w_{\vec x}$, in order to
derive $t$, it is enough to sum in an increasing way the negative
weights $w_i$ occurring in $\vec w_{\vec x}$ until this sum is lower
than or equal to the opposite of the sum of all the positive weights
occurring in $\vec w_{\vec x}$. This time, the term $t$ composed of
the variables $x_i$ corresponding to the negative weights that have
been selected is by construction a minimal sufficient reason for $\vec x$ given
$\vec w_{\vec x}$.



All the experiments have been conducted on a computer equipped with Intel(R) XEON E5-2637 CPU @ 3.5 GHz and 128 Gib of memory. 
A time-out (TO) of 600s has been considered for each instance and each type of explanation, except LIME explanations. 
\paragraph{Results.}
A first conclusion that can be drawn from our experiments is the intractability of computing in practice minimal sufficient reasons 
(this is not surprising, since this coheres with the complexity result given by Proposition \ref{prop:complexityminimalMUS}).
Indeed, we have been able to compute within the time limit of 600s a minimal reason for only 10 instances and a single
dataset (\textit{compas}). 

We report hereafter empirical results about two datasets only, namely \textit{placement} and \textit{gisette} (the results obtained on the other datasets are 
similar and available as at \url{http://www.cril.univ-artois.fr/expekctation/}). The \textit{placement} data set is about the placement of students in a  campus.  
It consists of 215 labelled instances. Students are described using 13 features, related to their curricula,
the type and work experience and the salary. An instance is labelled as positive when the student gets a job. 
The random forest that has been generated consists of 25 trees, and its mean accuracy is 97.6\%.
\textit{gisette} is a much larger dataset, based on 5000 features and containing 7000 labelled instances. 
Features correspond to pixels. The problem is to separate the highly confusible digits 4 and 9. An instance is labelled as positive  whenever the picture represents a 9.
The random forest that has been generated consists of 85 trees, and its mean accuracy is 96\%.

\begin{figure}[t]
  \centering
  \begin{subfigure}[t]{.45\linewidth}
    \includegraphics[width=\linewidth,height=4.0cm]{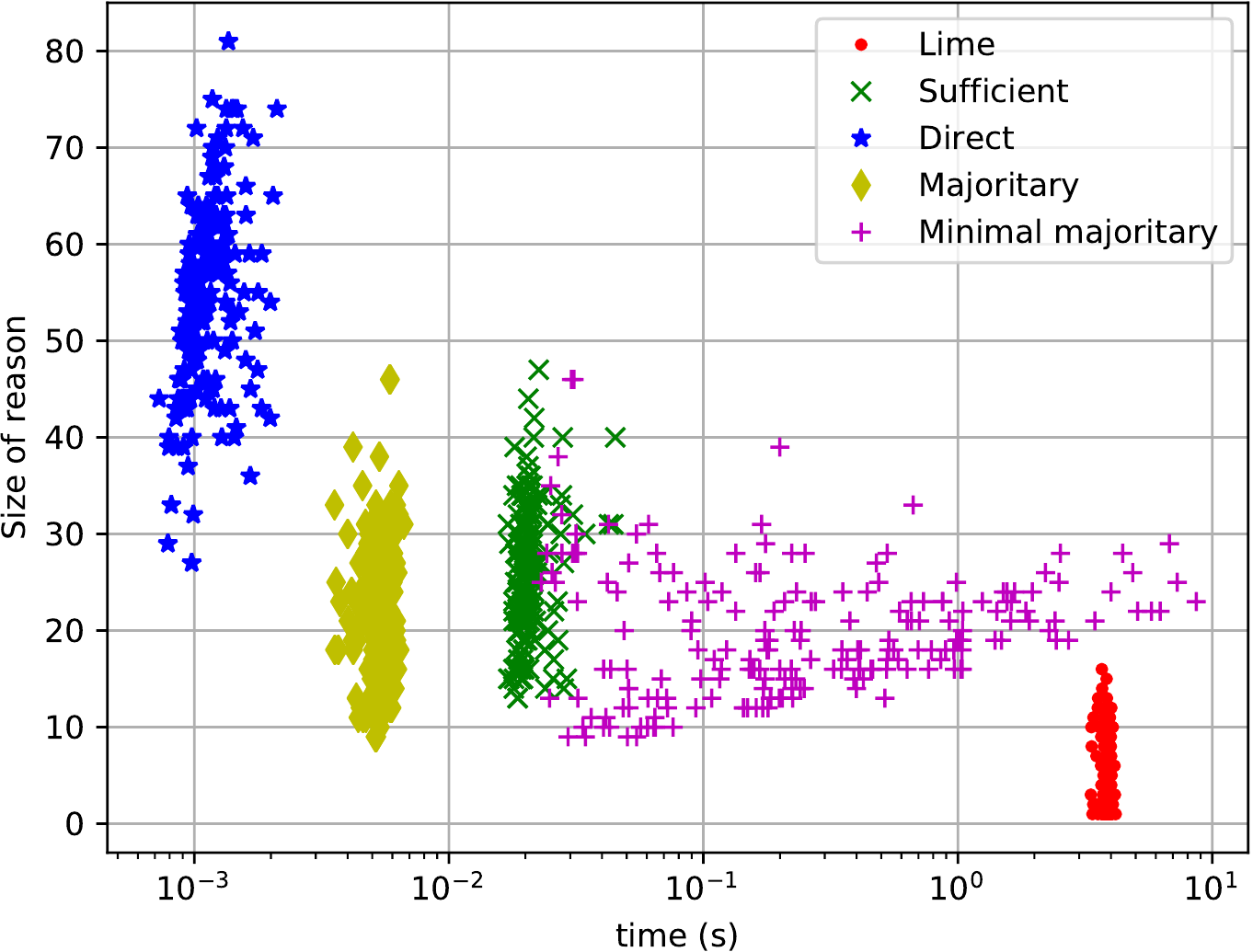}
\end{subfigure}
  \begin{subfigure}[t]{.45\linewidth}
    \includegraphics[width=\linewidth,height=4.0cm]{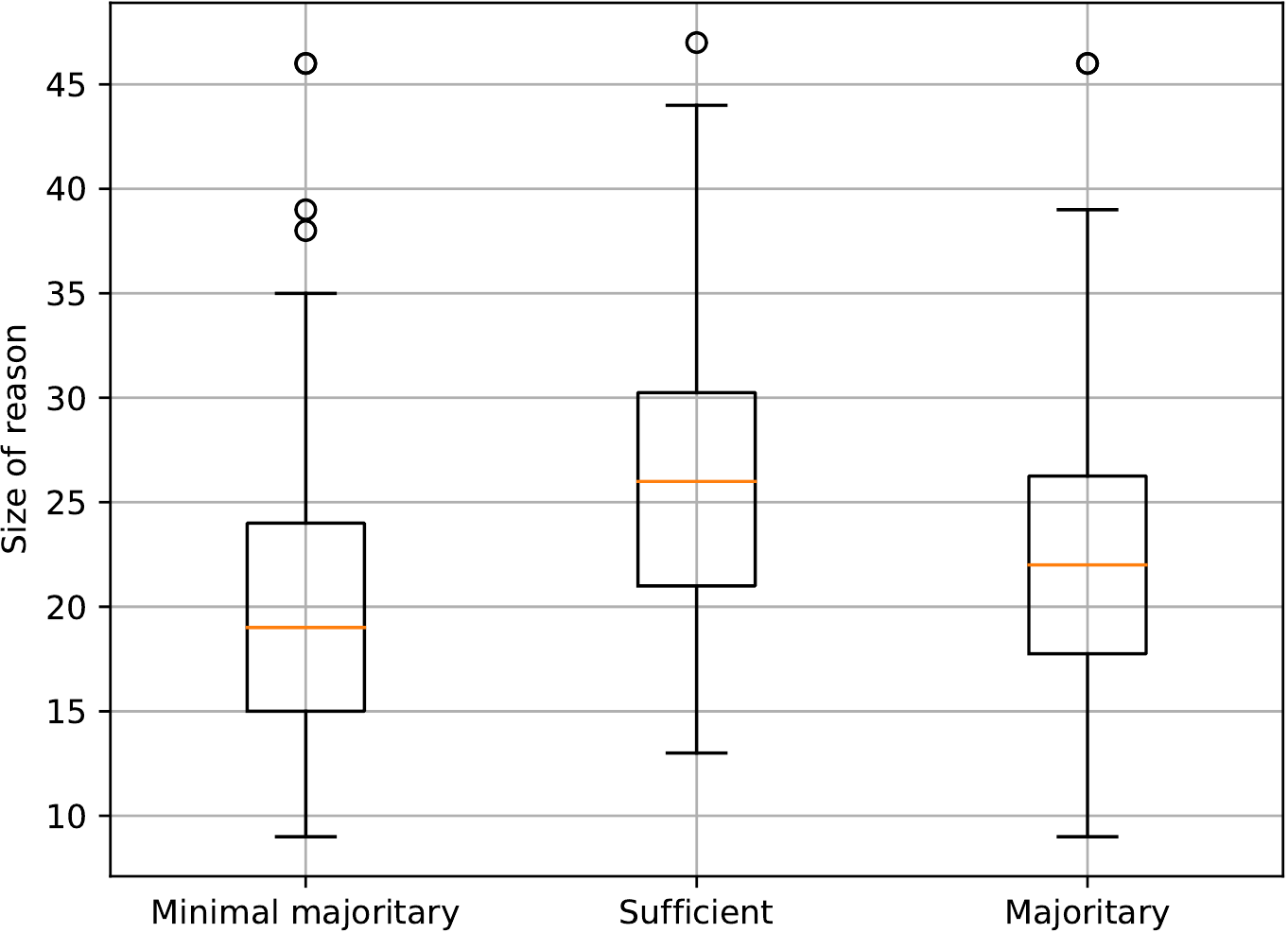}
    \vspace*{-0.2cm}
\end{subfigure}
  \begin{subfigure}[b]{.45\linewidth}
    \includegraphics[width=\linewidth,height=4.0cm]{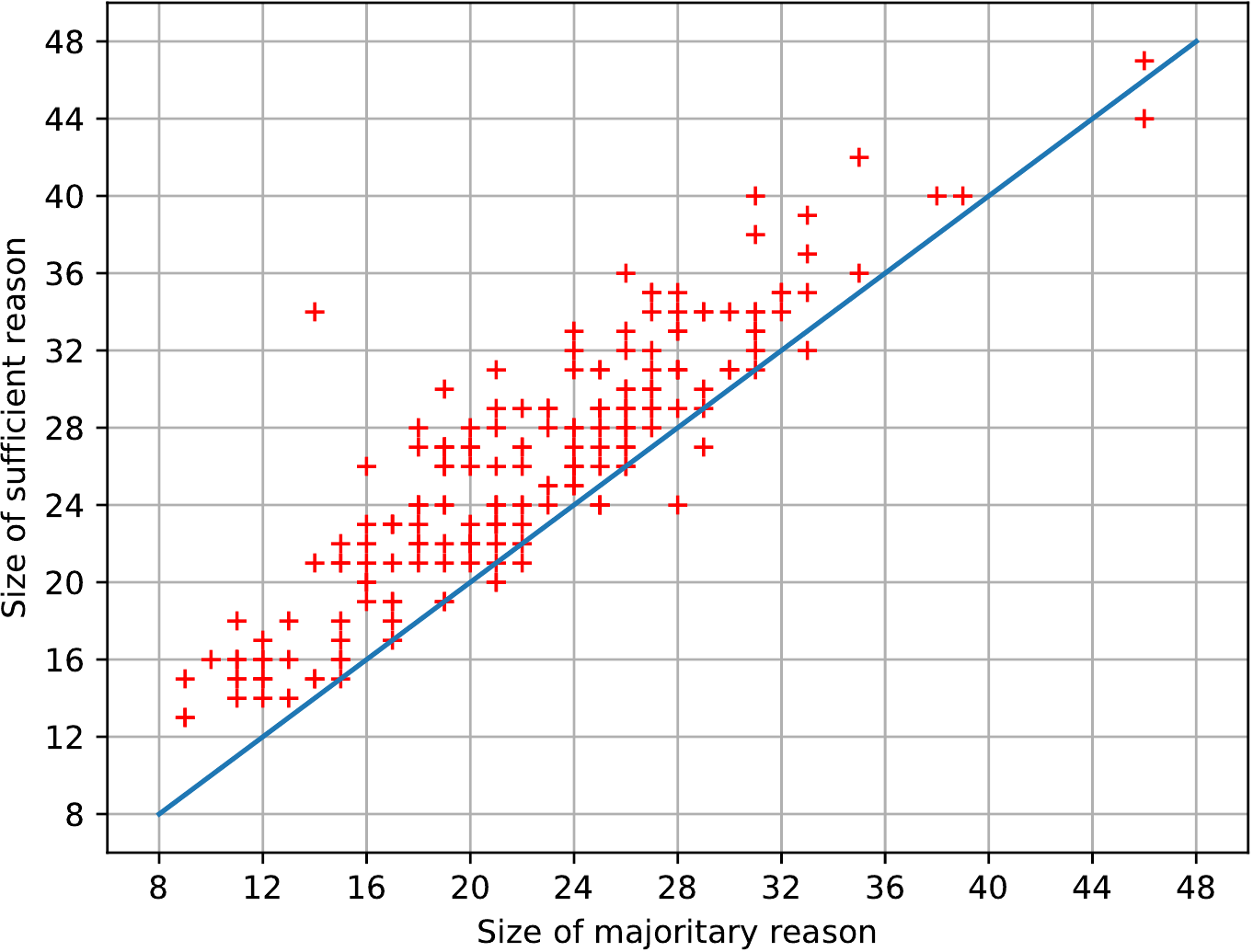}
\end{subfigure}
  \begin{subfigure}[b]{.45\linewidth}
    \includegraphics[width=\linewidth,height=4.0cm]{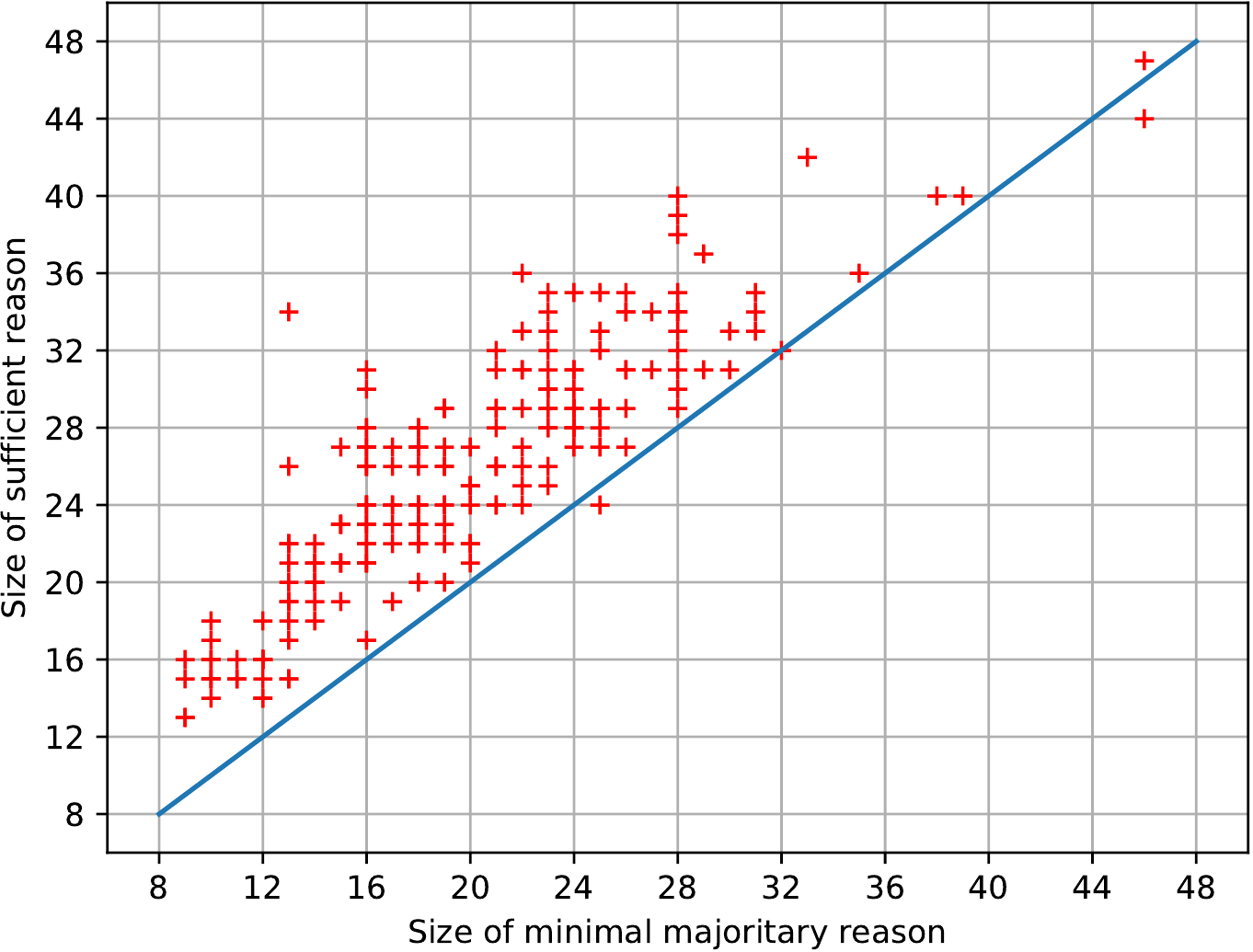}
\end{subfigure}
\caption{\label{fig:placement}Empirical results for the \textit{placement} dataset.}
\end{figure}

Figure \ref{fig:placement} provides the results obtained for \textit{placement}, using four plots.
Each dot represents an instance. The first plot shows the time needed to compute a reason on the x-axis, and 
the size of this reason on the y-axis. On this plot, no dot corresponds to a minimal sufficient reason because their computation did not terminate
before the time-out.  The plot also highlights that all the other reasons have been computed within the time limit, and in general using a small amount of time.
In particular, it shows that the direct reason can be quite large, that the computation of LIME explanations is usually more expensive than 
the ones of the other explanations, and that LIME explanations can be very short (but one must keep in mind that they are not 
abductive explanations in general\footnote{See also \cite{DBLP:conf/sat/NarodytskaSMIM19}
that reports some experiments about {\sc Anchor} (the successor of LIME), assessing the quality of the explanations computed using {\sc Anchor}.}).
A box plot about the sizes of all the explanations is reported (the LIME ones and the direct reasons are not presented for the sake of readibility). 
The figure also provides two scatter plots, aiming to compare the size of majoritary reasons with the size of sufficient reasons, as well as the 
size of the minimal majoritary reasons with the size of sufficient reasons. 
These plots clearly show the benefits that can be offered by considering majoritary reasons and minimal majoritary reasons instead of
sufficient reasons.

\begin{figure}
  \centering
  \begin{subfigure}[t]{.45\linewidth}
    \includegraphics[width=\linewidth,height=4.0cm]{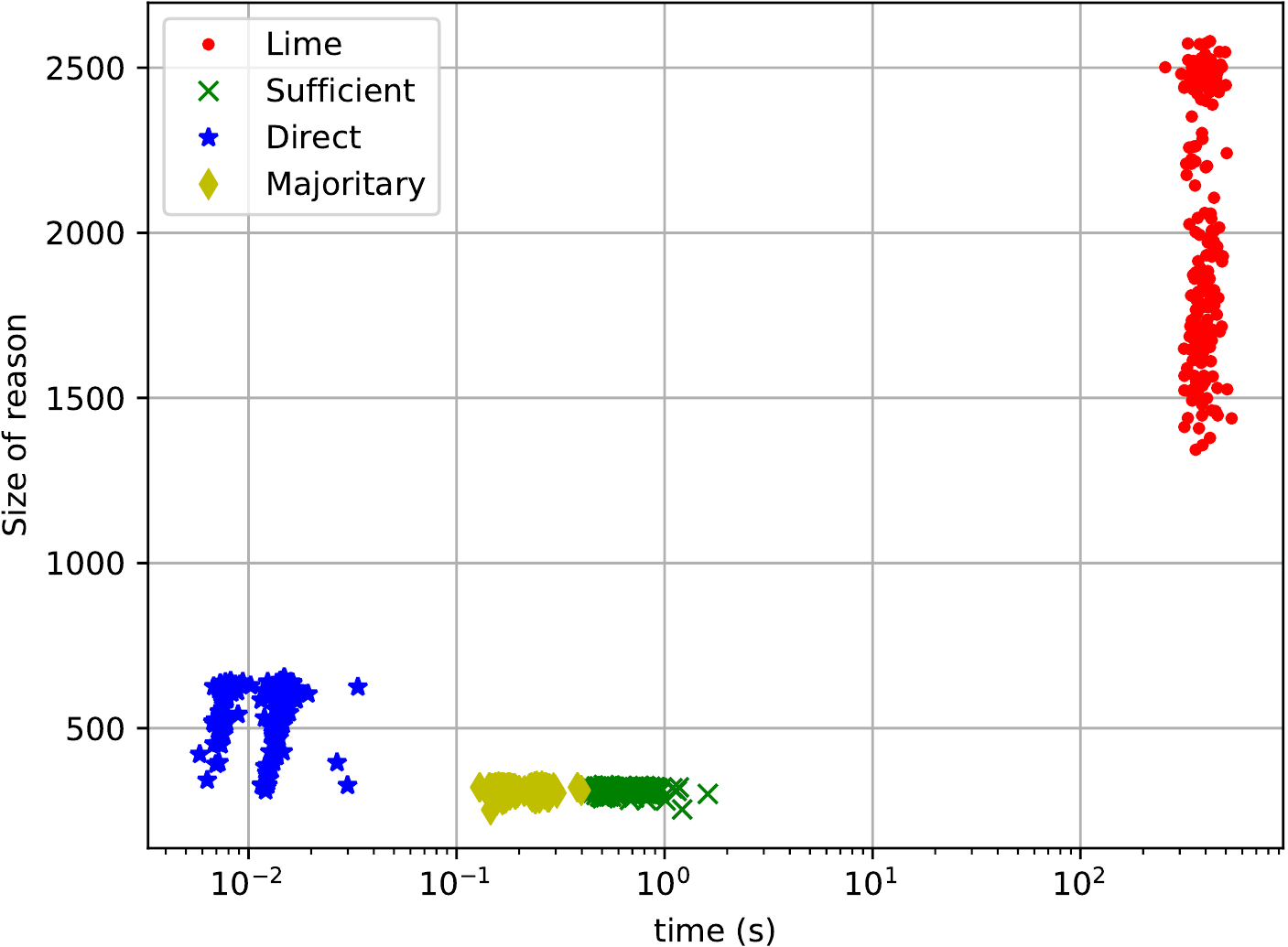}
\end{subfigure}
  \begin{subfigure}[t]{.45\linewidth}
    \includegraphics[width=\linewidth,height=4.0cm]{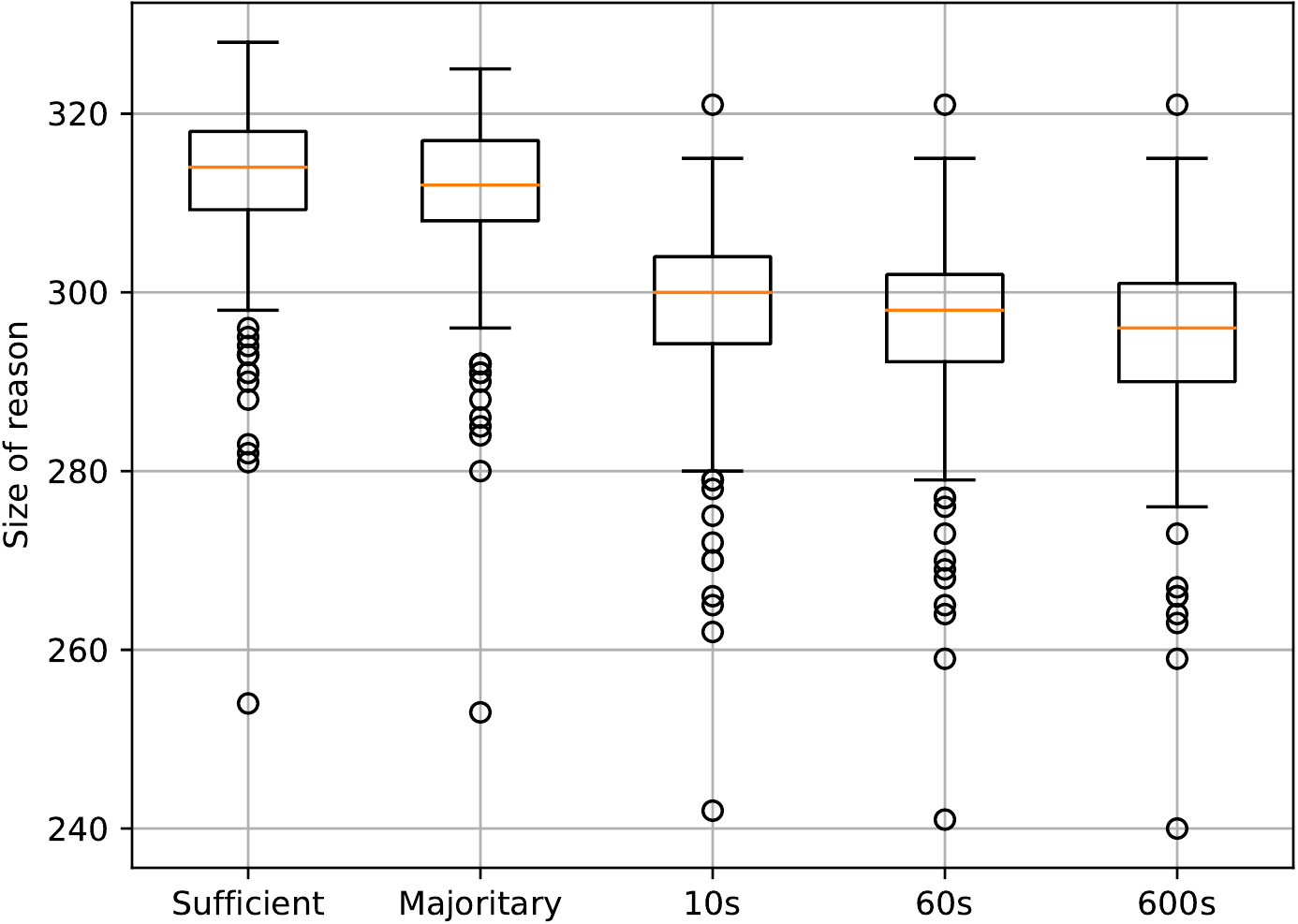}
    \vspace*{-0.2cm}
\end{subfigure}
  \begin{subfigure}[b]{.45\linewidth}
    \includegraphics[width=\linewidth,height=4.0cm]{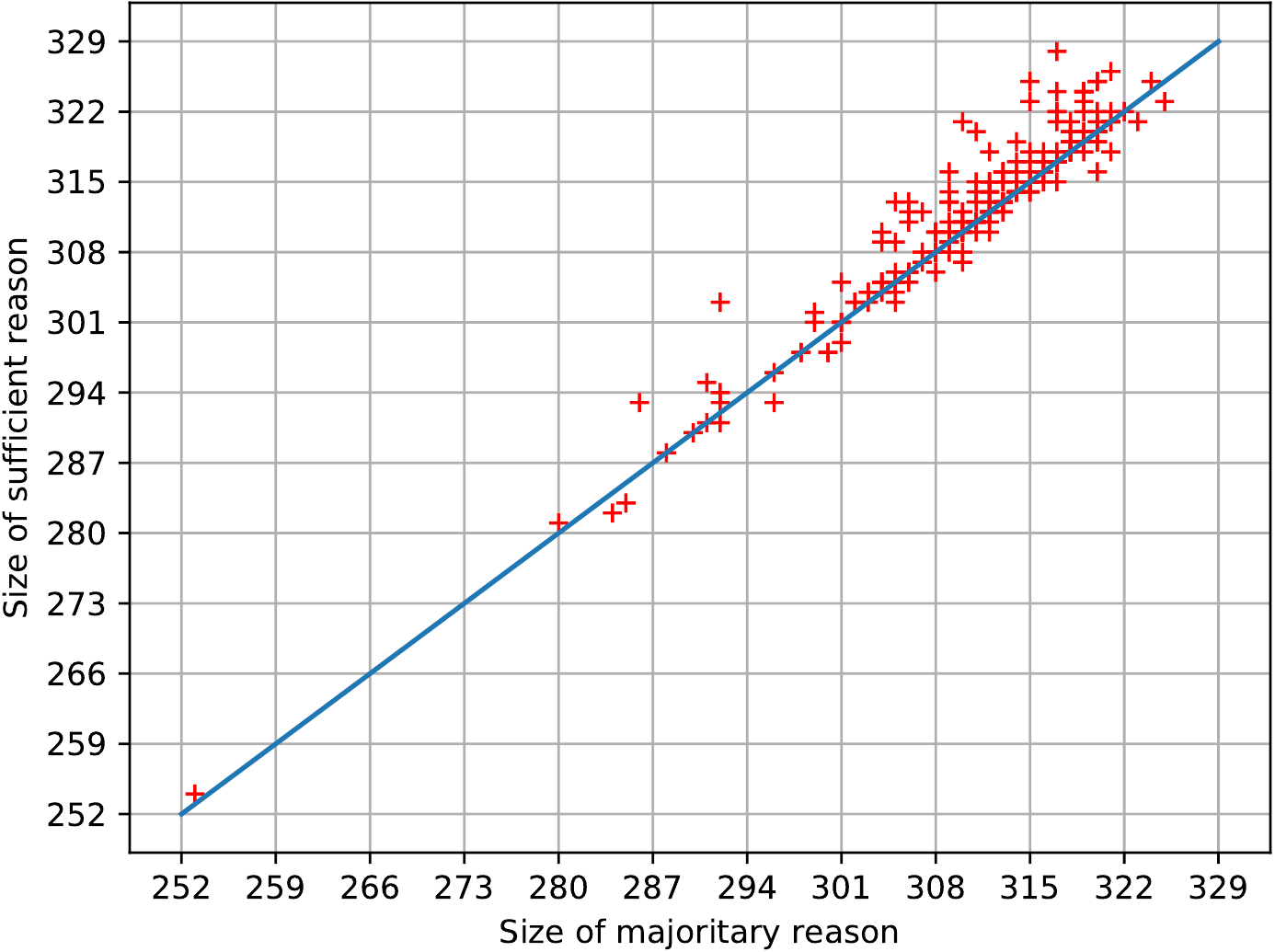}
\end{subfigure}
  \begin{subfigure}[b]{.45\linewidth}
    \includegraphics[width=\linewidth,height=4.0cm]{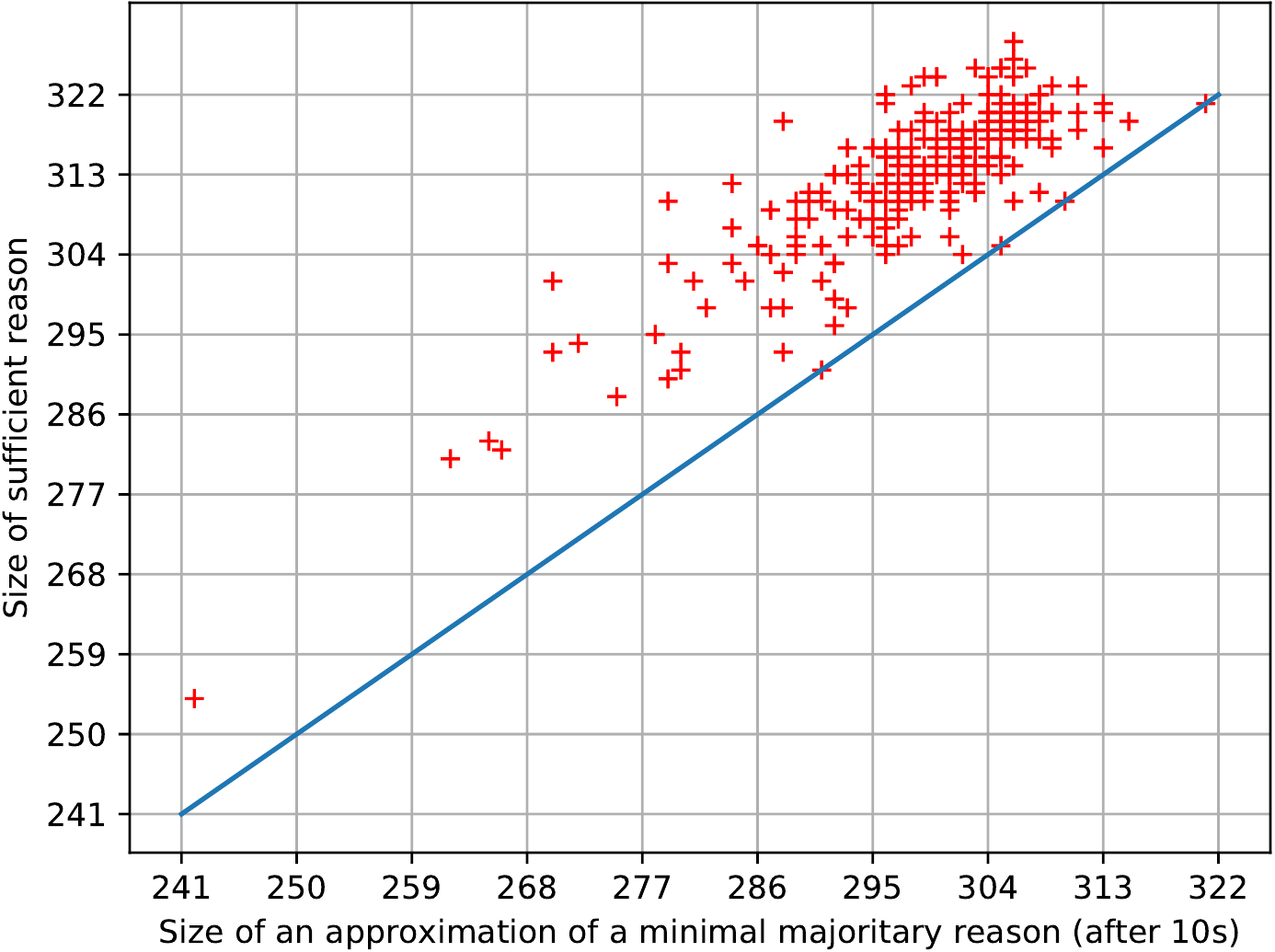}
\end{subfigure}
\caption{\label{fig:gisette}Empirical results for the \textit{gisette} dataset.}
\end{figure}

Figure \ref{fig:gisette} synthesizes the results obtained for \textit{gisette}, using four plots again.
Three of them are of the same kind as the plots used for \textit{placement}.
Conclusions similar to those drawn for \textit{placement} can be derived for \textit{gisette}, with some exceptions. 
First of all, this time, no dot corresponds to a minimal majoritary reason because their computation did not terminate before the time-out.  
Furthermore, LIME explanations are very long here. This can be explained by the fact that the computation achieved by LIME relies on a 
binary representation of the instance that is quite different (and possibly much larger) than the one considered in the representation 
of the random forest. Indeed, each decision tree of the forest focuses only on a subset of most important features (in the sense of Gini criterion) 
found during the learning phase. In our experiments, the size of LIME explanations was typically high for datasets based on many features.


When minimal majoritary reasons are hard to be computed (as it is the case for \textit{gisette}), an approach consists in approximating them.
Interestingly, one can take advantage of an incremental \textsc{partial MaxSAT algorithm}, 
like {\tt LMHS} \cite{SaikkoBJ16}, to do the job.
Specifically, the result given in Proposition \ref{prop:minimal-weightoptim} provides a way to derive abductive explanations 
for an instance $\vec{x}$ given a random forest $F$ in an \emph{anytime} fashion. 
Basically, using {\tt LMHS}, a Boolean assignment $\vec{z}$ satisfying all the hard constraints 
of $C_{\mathrm{hard}}$ and a given number, say $k$, of soft constraints from $C_{\mathrm{soft}}$ is looked for ($k$ is set to 0 at start). 
If such an assignment is found, then one looks for an assignment satisfying $k+1$ soft constraint, and so on, until an optimal solution is found 
or a preset time bound is reached. In many cases, the most demanding step from a computational standpoint is the one for which 
$k$ is the optimal value (but one ignores it) and one looks for an assignment  that satisfies $k+1$ soft constraint 
(and such an assignment does not exist). By construction, every $\vec{z}$ that is generated that way is such that 
$t_{\vec x} \cap t_{\vec z}$ is an implicant of $F$ that covers $\vec {x}$ (and hence, an abductive explanation).
The approximation $\vec{z}$ of a minimal majoritary reason for $\vec{x}$ given $F$, which is obtained when the time limit is met, can be 
significantly shorter than the sufficient reason for $\vec{x}$ given $F$ that has been derived.  
In our experiments, we used three time limits: 10s, 60s, 600s. 
As the box plot and the dedicated scatter plot given in Figure \ref{fig:gisette} show it, the sizes of the approximations $\vec{z}$ which are derived 
gently decrease with time. Interestingly, the size savings that are achieved in comparison to sufficient reasons are significant, 
even for the smallest time bound of 10s that has been considered.


\section{Conclusion}\label{conclusion}

In this paper, we have introduced, analyzed and evaluated some new notions of abductive explanations suited to random forest classifiers, namely
majoritary reasons and minimal majoritary reasons. 
Our investigation reveals the existence of a trade-off between runtime complexity and sparsity for abductive explanations.
Unlike sufficient reasons, majoritary reasons and minimal majoritary reasons may contain irrelevant
features. Despite this evidence,  majoritary reasons and minimal majoritary reasons appear as valuable alternative to sufficient reasons. 
Indeed, majoritary reasons can be computed in polynomial time while sufficient reasons cannot (unless {\sf P = NP}).
In addition, most of the time in our experiments, majoritary reasons appear as slightly smaller than sufficient reasons. 
Minimal majoritary reasons can be looked for when majoritary reasons are too large, but this is at the cost of an extra computation time 
that can be important, and even prohibitive in some cases. However, minimal majoritary reasons can be approximated using
an \emph{anytime}  \textsc{Partial MaxSAT} algorithm. Empirically, approximations can be derived within a small amount of time
and their sizes are significantly smaller than the ones of sufficient reasons.

\section*{Acknowledgements}
This work has benefited from the support of the AI Chair EXPE\textcolor{orange}{KC}TATION (ANR-19-CHIA-0005-01) of the French National Research Agency.
It was also partially supported by TAILOR, a project funded by EU Horizon 2020 research
and innovation programme under GA No 952215.

\bibliographystyle{plain}
\bibliography{biblio}

\section*{Proofs}

For every $f, g \in \mathcal F_n$, we note $f \models g$ when for every $\vec x \in \{0,1\}^n$, $f(\vec x) = 1$ implies that $g(\vec x) = 1$.

\medskip

\noindent {\bf Proof of Proposition \ref{prop:transfoRF}}

\begin{proof}
By definition, an instance $\vec x$ is a model of the negation of a given random forest $F = \{T_1, \ldots, T_m\}$ if and only if
it is a model of at most $\frac{m}{2}$ trees among those of $T$. Let us state (w.l.o.g.) that $\vec x$ is a model of $T_1, \ldots, T_k$ and
a counter-model of $T_{k+1}, \ldots, T_m$, with $k \leq \frac{m}{2}$. Equivalently, we have that $\vec x$ is a counter-model of $T'_1, \ldots, T'_k$ and
a model of $T'_{k+1}, \ldots, T'_m$ where each $T'_i$ ($i \in \{1, \ldots, m\}$) is a decision tree equivalent to the negation of $T_i$.
This precisely means that $\vec x$ is a model of $\neg F = \{T'_1, \ldots, T'_m\}$. Since each $T'_i$ ($i \in \{1, \ldots, m\}$) can be computed in time linear 
in $\size{T_i}$, the result follows.
\end{proof}

\noindent {\bf Proof of Proposition \ref{prop:CNF2RF}}

\begin{proof}
    Let $G = c_1 \land \cdots \land c_p$ be a \cnf\  formula with $p > 0$ clauses. 
    Each $c_i$ can be transformed into a decision tree $T_i$ using the following linear-time recursive algorithm. For the base cases, 
    if $c_i$ is empty, then $T_i = 0$ and if $c_i$ is a tautology, then $T_i = 1$. For the inductive case, suppose that $c_i = l_j \lor c'_i$ and 
    let $T'_j$ the decision tree encoding $c'_i$. If $l_j = x_j$ (resp. $l_j = \overline x_j$), then $T_i$ is the decision tree rooted at $x_j$ with right (resp. left) 
    child labeled by the leaf $1$ (resp. $0$) and with left (resp. right) child encoding $T'_j$. 
    Now, let $F = \{T_1, \cdots, T_p, T_{p+1}, \cdots, T_{q}\}$, where $T_{p+1},\cdots, T_q$ are decision trees rooted at $0$, and $q = 2p - 1$. 
    For any input instance $\vec x \in \{0,1\}^n$, we have $G(\vec x) = 1$ iff $T_i(\vec x) = 1$ for every $i \in [p]$.
    Since for any positive integer $p$, the function 
    $$\phi_p(z) = \frac{p - z}{2p - 1}$$ 
    satisfies $\phi_p(0) > \nicefrac{1}{2}$ and $\phi_p(z) < \nicefrac{1}{2}$ for $z \in [p]$,
    it follows that $G(\vec x) = 1$ iff $F(\vec x) > \nicefrac{1}{2}$.
    
    The case for \dnf\ formulas is dual: given $G = t_1 \lor \cdots \lor t_p$, compute in linear time a \cnf\ formula equivalent to $\neg G$, then turn it in linear time into an equivalent
    random forest using the transformation above, and finally negate in linear time the resulting random forest by taking advantage of Proposition~\ref{prop:transfoRF}.  
    
   Since every \cnf\ formula $G$ can be turned in linear time into an equivalent random forest $G'$ (as we have just proved it), the size of $G'$ is polynomial
     in the size of $G$ for a fixed polynomial (independent of $G$). Then, exploiting Proposition \ref{prop:transfoRF}, one can negate $G'$ in linear time.
     The resulting random forest $G''$ is equivalent to the negation of $G$ and its size is also polynomial in the size of $G$ for a fixed polynomial.
     
     Finally, suppose, towards a contradiction, that a polynomial-space translation from \rf\ to \cnf\ exists. If so, one could compute a \cnf\ formula $G'''$ equivalent to $G''$
     and having a size polynomial in the size of $G''$ for a fixed polynomial. Thus, the \cnf\ formula $G'''$ would have a size polynomial in the size of $G$ for a fixed polynomial.
     This \cnf\ formula could be negated in linear time into a \dnf\ formula $G''''$ by applying De Morgan's laws. By construction, $G''''$ would be a \dnf\ formula equivalent to
     $G$, and its size would be polynomial in the size of $G$ for a fixed polynomial. This conflicts with the fact that there is no polynomial-space translation from \cnf\ to \dnf, see e.g., \cite{DarwicheMarquis02}. Using duality, we prove similarly that there is no polynomial-space translation from \rf\ to \dnf.
     
    \end{proof}

\noindent {\bf Proof of Proposition \ref{prop:complexityimplicanttestRF}}
\begin{proof}~
\begin{itemize}
\item Membership to {\sf coNP}: we show that the complementary problem, i.e., the problem of deciding whether a term $t$ is not an implicant of 
a random forest $F$, is in {\sf NP}. This is direct given the characterization of the implicants of $F$ provided by Proposition \ref{prop:implicanttestRF}: 
it is enough to comput  in time polynomial in the size of $F$ the \cnf\ formula $H$ given in Proposition \ref{prop:implicanttestRF}, and to exploit the fact that 
$t$ is not an implicant of $F$ if and only if $t \wedge H$ is satisfiable. Finally, 
deciding whether $t \wedge H$ is satisfiable can be easily achieved by a non-deterministic algorithm running in time polynomial in the size of the input 
(just guess a truth assignment over the variables occurring in $t \wedge H$ and check in polynomial time that this assignment is a model of $t \wedge H$).

\item {\sf coNP}-hardness: by reduction from {\sc VAL}, the validity problem for \dnf\ formulae. Let $G = d_1 \vee \ldots \vee d_p$ be a \dnf\ formula over $X_n$. We can associate with $G$ in polynomial time an equivalent random forest $F$ using Proposition \ref{prop:CNF2RF}. Consider now the term $t = \top$. $t$ is an implicant of $F$ if and only if $G$ is valid.
\end{itemize}
\end{proof}

\noindent {\bf Proof of Proposition \ref{prop:implicanttestRF}}    
\begin{proof}
We have $t \not \models F$ if and only if $t \wedge \neg F$ is satisfiable.
From $F$, exploiting Proposition \ref{prop:transfoRF} one can generate in polynomial time a random forest $F' = \{\neg T_1, \ldots, \neg T_m\}$ equivalent to $\neg F$.
Each $\neg T_i$ is the decision tree obtained by replacing every $1$-leaf in $T_i$ by a $0$-leaf, and vice-versa.
We thus have $t \not \models F$ if and only if $t \wedge F'$ is satisfiable. Then $F'$ can be associated in polynomial time with 
the following Boolean quantified formula $\exists Y . H$ when $Y = \{y_i : i \in [m]\} \cup A$ is a set of new variables and 
$H$ is a \cnf\ formula which is the conjunction of the clauses of  $$\{(\overline y_i \vee c) : i \in [m], c \in \cnf(T'_i) \}$$ 
with a \cnf\ encoding of the cardinality constraint $$\sum_{i = 1}^m y_i > \frac{m}{2}$$ using
auxiliary variables in $A$. $F'$ is equivalent to  $\exists Y . H$, therefore $t \wedge F'$ is satisfiable if and only if 
$t \wedge \exists Y . H$ is satisfiable. Since the variables of $Y$ do not occur in $t$, $t \wedge \exists Y . H$ is equivalent to
$\exists Y . (t \wedge H)$. Since $\exists Y . (t \wedge H)$ is satisfiable if and only if $t \wedge H$ is satisfiable, we get that
$t \wedge \exists Y . H$ is satisfiable if and only if $t \wedge H$ is satisfiable. 
\end{proof}

\noindent {\bf Proof of Proposition \ref{prop:complexityminimalMUS}}

\begin{proof}~
\begin{itemize}
\item Membership to $\Sigma_2^p$: if there exists a minimal reason $t$ for $\vec x$ given $f$ such that $t$ contains at most $k$ features, then one can 
guess $t$ using a nondeterministic algorithm running in polynomial time (the size of $t$ is bounded by the size of $\vec x$), then check in polynomial time that $t$ is a 
sufficient reason for $\vec x$ given $f$ using an {\sf NP}-oracle (this comes directly from the fact that this problem belongs to {\sf DP}),  
and finally check in polynomial time that the size of $t$ is upper bounded by $k$.
\item $\Sigma_2^p$-hardness: in \cite{DBLP:journals/ai/Liberatore05} (Theorem 2), it is shown that the problem of deciding whether a \cnf\ formula $\Pi = \bigwedge_{i=1}^p c_i$ 
has an irredundant equivalent subset of size at most $k$ is $\Sigma_2^p$-complete, and that the problem is $\Sigma_2^p$-hard 
even in the case when $\Pi$ is unsatisfiable. Whenever $\Pi$ is unsatisfiable, an irredundant equivalent subset of $\Pi$ precisely is a MUS of $\Pi$ 
(every clause being considered as a soft clause). Accordingly, there exists an irredundant equivalent subset $E$ of an unsatisfiable \cnf\ formula $\Pi$ such $E$ is of size at most $k$
if and only if there exists a MUS $I = \{y_i : c_i \in E\}$ of $S = \{y_i : c_i \in \Pi\}$ given $H = \{\overline{y}_i \vee c_i : c_i \in \Pi\}$ such that $I$ is of size at most $k$.
Because of this equivalence, the problem of deciding whether $S$ has a MUS of size at most $k$ given $H$ has the same complexity as the problem
of  deciding whether $\Pi$ has an irredundant equivalent subset of size at most $k$, so it is $\Sigma_2^p$-hard.
Finally, we reduce this latter problem to the one of deciding whether a term is a minimal reason for an instance given a random forest. 
The reduction is as follows. With $(H, S)$ where $S$ is satisfiable and $H \cup S$ is unsatisfiable (as obtained from the previous reduction),
one associates in polynomial time the pair $(\vec x, F)$ where $\vec x$ is any interpretation that extends $S$ and
$F$ is a random forest from $\rf_{n, m}$ equivalent to $\neg H$ (since $H$ is a \cnf\ formula, a \dnf\ formula equivalent to $\neg H$ can be computed in linear time from $H$ and turned
in linear time into an equivalent random forest $F$ as shown by Proposition \ref{prop:CNF2RF}). Since $H \cup S$ is unsatisfiable, we have $S \models \neg H$ showing
that $\vec x \models F$. Now, $I$ is a MUS of $S$ given $H$ if and only if $I \cup H$ is unsatisfiable and for every $l \in I$, $(I \setminus \{l\}) \cup H$ is satisfiable.
Taking $t = I$, this is equivalent to state that $t \wedge \neg F$ is unsatisfiable and for every $l \in t$, $(t \setminus \{l\}) \wedge \neg F$ is satisfiable.
Equivalently, $t \models F$ and for every $l \in t$, $(t \setminus \{l\}) \not \models F$, or stated otherwise $t$ is a prime implicant of $F$. 
Since $t = I$ and $I \subseteq S$, we also have $S \models t$, hence $\vec x \models t$.  
Thus $t$ is a sufficient reason for $\vec x$ given $F$. Since $|I| = |t|$, a MUS $I$ of $S$ given $H$ such that $|I| \leq k$ exists if and only if
a sufficient reason $t$ for $\vec x$ given $F$ such that $|t| \leq k$ exists. This completes the proof.
\end{itemize}
\end{proof}

\noindent {\bf Proof of Proposition \ref{prop:majoritary}}
\begin{proof}
Again, we focus only on the case when $F(\vec x) = 1$ (if $F(\vec x) = 0$, it is enough to consider the random forest $\neg F$ instead of $F$). 

If $F$ contains at most $2$ trees, then $F$ is equivalent to the conjunction of its elements. In this case, testing whether a term $t$ implied by $\vec x$ is an implicant
of $F$ boils down to testing that $t$ is an implicant of every tree in $F$, so that the sufficient reasons for $\vec x$ given $F$ are precisely the majoritary reasons 
for $\vec x$ given $F$.

As to the case $m \geq 3$, whatever $n \geq 1$, let $T$ be a decision tree equivalent to the parity function $\oplus_{i=1}^n x_i$. Consider the random forest $F$ containing $\lfloor \frac{m}{2} \rfloor$ copies of $T$,
$\lfloor \frac{m}{2} \rfloor$ copies of the decision tree $\neg T$, and a decision tree reduced to a $1$-leaf. By construction, $F$ is valid. Indeed, among the subsets of $F$ containing a strict
majority of decision trees, one can find the one containing all the $\lfloor \frac{m}{2} \rfloor$ copies of $T$ plus the $1$-leaf (their conjunction is thus equivalent to $T$) and
the one containing all the $\lfloor \frac{m}{2} \rfloor$ copies of $\neg T$ plus the $1$-leaf (their conjunction is thus equivalent to $\neg T$). Their disjunction is thus valid.
As a consequence, whatever $\vec x$, we have $F(\vec x) = 1$. Indeed, we have either $T(\vec x) = 1$ or $\neg T(\vec x) = 1$ (and obviously, $1(\vec x) = 1$). Thus,
$t_{\vec x}$ is an implicant of a (strict) majority of decision trees of $F$.
Now, consider any literal $l$ of $t_{\vec x}$. The term $t _{\vec x} \setminus \{l\}$ is not an implicant of $T$ nor an implicant of $\neg T$ since the implicants of the parity function
$\oplus_{i=1}^n x_i$ (or of its negation) depend on every variable $x_i$ ($i \in \{1, \ldots, n\}$). Therefore, $t_{\vec x}$ is the unique majoritary reason for
$\vec x$ given $F$ and it contains $n$ characteristics. But since $F$ is valid, $\top$ is the unique sufficient reason for $\vec x$ given $F$.
\end{proof}

\noindent {\bf Proof of Proposition \ref{prop:minimal majoritaryreasonRF}}
\begin{proof}~
\begin{itemize}
\item Membership to {\sf NP}:  if there exists a minimal majoritary reason $t$ for $\vec x$ given $F$ such that $t$ contains at most $k$ features, then one can 
guess $t$ using a nondeterministic algorithm running in polynomial time (the size of $t$ is bounded by the size of $\vec x$), then check in polynomial time whether $t$ is a 
sufficient reason for $\vec x$ given $T$ for a majority of trees $T \in F$, and finally check in polynomial time that the size of $t$ is upper bounded by $k$.

\item {\sf NP}-hardness:  in the following, we focus only on the case when $F(\vec x) = 1$ (if $F(\vec x) = 0$, it is enough to consider the random forest $\neg F$ instead of $F$; 
this is harmless given that $\neg F$ can be computed in time linear in the size of $F$, see Proposition \ref{prop:transfoRF}). We assume that $m=1$, i.e., $F$
consists of a single decision tree $T \in \dt_n$.

We call \textsc{Minimal Sufficient Reason} the problem that asks, given $T \in \dt_n$, $\vec x \in \{0,1\}^n$ with $T(x) = 1$ and $k \in \mathbb N$,
    whether there is an implicant $t$ of $T$ of size at most $k$ that covers $\vec x$.  
%
    
    Our objective is to prove that \textsc{Minimal Sufficient Reason} is {\sf NP}-hard. 
    To this end, let us first recall that a \emph{vertex cover} of an undirected graph $G = (X,E)$
    is a subset $V \subseteq X$ of vertices such that $\{y,z\} \cap V \neq \emptyset$ for 
    every edge $e = \{y,z\}$ in $E$. In the \textsc{Min Vertex Cover} problem, we are given a graph $G$ together with an integer 
    $k \in \mathbb N$, and the task is to find a vertex cover $V$ of $G$ of size at most $k$. 
    \textsc{Min Vertex Cover} is a well-known {\sf NP}-hard problem \cite{Karp1972}, 
    and we now show that it can be reduced in polynomial time to \textsc{Minimal Sufficient Reason}.

    Suppose that we are given a graph $G = (X,E)$ and assume, without loss of generality, that $G$ does not include isolated vertices.  
    For any $y \in X$, let $E_y = \{e \in E: y \in e\}$ denote the set of edges in $G$ that are adjacent to $y$, 
    and let $N_y = \{z \in X: \{y,z\} \in E\}$ denote the set of neighbors of $y$ in $G$.
    By $G \setminus y$, we denote the deletion of $y$ from $G$, obtained by removing $y$ and its adjacent edges, 
    i.e., $G \setminus y = (X \setminus \{y\}, E \setminus E_y)$. 
    We associate with $G$ a decision tree $T(G)$ over $X_n = X$ using the following recursive algorithm.
    If $G$ is the empty graph (i.e. $E = \emptyset$), then return the decision tree rooted at a $1$-leaf.
    Otherwise, pick a node $y \in X$ and generate a decision tree $T(G)$ such that:
    \begin{enumerate}
        \item[(1)] the root is labeled by $y$;
        \item[(2)] the left child is the decision tree encoding the monomial $\bigwedge N_y$;
        \item[(3)] the right child is the decision tree $T(G')$ returned by calling the algorithm on $G' = G \setminus y$.
    \end{enumerate}
    By construction, $T(G)$ is a complete backtrack search tree of the formula 
    $\cnf(E) = \bigwedge \{(y \lor z): \{y,z\} \in E\}$, 
    which implies that $T(G)$ and $\cnf(E)$ are logically equivalent.
    Furthermore, $T(G)$ is a comb-shaped tree since recursion only on the rightmost branch. 
    In particular, the algorithm runs in $\mathcal O(n \size{E})$ time, since step (1) takes $\mathcal O(1)$ time, step (2) takes 
    $\mathcal O(n)$ time, and step (3) is called at most $\size{E}$ times. 
    
    Now, with an instance $\vec P_1 = (G,k)$ of \textsc{Min Vertex Cover}, 
    we associate the instance $\vec P_2 = (T(G),\vec x,k)$ of \textsc{Minimal Sufficient Reason}, where $\vec x = (1,\cdots,1)$. 
    Based on the above algorithm, $\vec P_2$ can be constructed in time polynomial in the size of $\vec P_1$.
    
    Let $V$ be a solution of $\vec P_1$. Since $V$ is a vertex cover of $G$, the term $t_V = \bigwedge V$ 
    is an implicant of the formula $\cnf(E)$.
    Since $t_V \subseteq t_{\vec x}$ and $\size{t_V} \leq k$, it follows from the fact that $\cnf(E)$ and $T(G)$ are 
    logically equivalent that $t_V$ is a solution of $\vec P_2$.  

    Conversely, let $t$ be a solution of $\vec P_2$. Since $t$ is an implicant of $T(G)$, it follows that $t$ is an implicant 
    of $\cnf(E)$. This together with the fact that $t \subseteq t_{\vec x}$ implies that the subset of vertices 
    $V \subseteq X_n$, satisfying $\bigwedge V = t$, is a vertex cover of $G$. Since $\size V \leq k$, it is therefore a solution of $\vec P_1$.
    \end{itemize}
\end{proof}    
    
\noindent {\bf Proof of Proposition \ref{prop:minimal-weightoptim}}
\begin{proof}

Let us first recall that the forgetting $\exists V . f$ of a set of variables $V$ in a formula $f$ denotes a 
formula that is a most general consequence of $f$ that is independent of $V$ (in the sense that it
is equivalent to a formula where no variable from $V$ occurs) \cite{LLM03}.

Let $\vec z^*$ be any optimal solution of $(C_{\mathrm{soft}}, C_{\mathrm{hard}})$.
On the one hand,  $\vec z^*$ is a model of $C_{\mathrm{hard}}$.
Let $V$ be the set of variables occurring in $C_{\mathrm{hard}}$ but not in $X_n$. Since $\vec z^* \models C_{\mathrm{hard}}$, we have that
$\exists V . \vec z^* \models \exists V . C_{\mathrm{hard}}$ (see \cite{LLM03}). Stated otherwise, the projection $\exists V . \vec z^*$ of $\vec z^*$ on $X_n$ implies the projection 
of $C_{\mathrm{hard}}$ on $X_n$. 

On the other hand, by construction, a consistent term $t$ over $X_n$ implies the projection 
of $C_{\mathrm{hard}}$ on $X_n$ if and only if $t$ is an implicant of more than $\frac{m}{2}$ decision trees of $F$. 
Thus, the term $\exists V . \vec z^*$ is an implicant of more than $\frac{m}{2}$ decision trees of $F$.

Finally, if $\vec z^*$ is an optimal solution of $(C_{\mathrm{soft}}, C_{\mathrm{hard}})$, then $\vec z^*$ satisfies a maximal number of soft clauses
from $C_{\mathrm{soft}}$. Since those soft clauses are precisely the negations of the literals occurring in $t_{\vec x}$, the term $t_{\vec z^*} \cap t_{\vec x}$
obtained from $\exists V . \vec z^*$ by removing every literal that coincides with a soft clause is still an implicant of more than $\frac{m}{2}$ decision trees of $F$.
Indeed, $C_{\mathrm{hard}}$ is monotone on $X_n$ and the polarity of every variable of $X_n$ in $C_{\mathrm{hard}}$ is the same as its polarity 
in $t_{\vec x}$. Since $\vec z^*$ satisfies a maximal number of soft clauses,
$t_{\vec z^*} \cap t_{\vec x}$ contains a minimal number of literals. As $t_{\vec z^*} \cap t_{\vec x} \subseteq t_{\vec x}$, $t_{\vec z^*} \cap t_{\vec x}$ 
is a minimal majoritary reason for $\vec x$ given $F$.
\end{proof}




\end{document}